\documentclass[letterpaper, 10 pt, conference]{ieeeconf}  

\IEEEoverridecommandlockouts                              

\usepackage{cite}

\usepackage{graphicx}

\usepackage[ruled,vlined]{algorithm2e}   
\usepackage{multirow}   
\usepackage{booktabs}
\usepackage{float}
\usepackage{amsmath}
\usepackage{url}
\usepackage{hyperref}

\title{\LARGE \bf
RF-Modulated Adaptive Communication Improves Multi-Agent Robotic Exploration
}

\author{Lorin Achey, Breanne Crockett, Christoffer Heckman, Bradley Hayes
\thanks{*This work was supported by the University of Colorado Boulder}
\thanks{All authors are with the University of Colorado Boulder, Boulder, CO 80309, USA
        {\tt\small \{lorin.achey, breanne.crockett, christoffer.heckman, bradley.hayes\}@colorado.edu}}%
}

\begin{document}

\maketitle
\thispagestyle{empty}
\pagestyle{empty}

\begin{abstract}
Reliable coordination and efficient communication are critical challenges for multi-agent robotic exploration of environments where communication is limited. This work introduces Adaptive-RF Transmission (ART), a novel communication-aware planning algorithm that dynamically modulates transmission location based on signal strength and data payload size, enabling heterogeneous robot teams to share information efficiently without unnecessary backtracking. We further explore an extension to this approach called ART-SST, which enforces signal strength thresholds for high-fidelity data delivery. Through over 480 simulations across three cave-inspired environments, ART consistently outperforms existing strategies, including full rendezvous and minimum-signal heuristic approaches, achieving up to a 58\% reduction in distance traveled and up to 52\% faster exploration times compared to baseline methods. These results demonstrate that adaptive, payload-aware communication significantly improves coverage efficiency and mission speed in complex, communication-constrained environments, offering a promising foundation for future planetary exploration and search-and-rescue missions.
\end{abstract}

\section{INTRODUCTION}

Robotic systems capable of dynamically adapting to uncertain and evolving environments are essential to autonomous exploration missions. Planetary exploration \cite{StOnge-PlanetaryExplorationRobotTeams2020, rockenbauer2024traversingmars}, search and rescue \cite{scherer2015autonomous-s&r-multiuav}, reconnaissance and surveillance \cite{hougen2000miniature} are applications which frequently demand a high level of coverage of the exploration area. To provide complete area coverage while minimizing the exploration time, autonomous systems must be capable of managing their resources efficiently during exploration.

Traditional single-robot systems face significant limitations in adaptability and efficiency when operating in remote and unpredictable landscapes. A single robot system introduces a single point of failure for a mission. This design choice leads to conservative mission planning, which can limit the exploration potential of the platform. In search and rescue scenarios, where coverage and response time are critical \cite{pyla2023design}, a single robot failure can result in loss of life. To address these challenges, multi-agent robotic systems that work collaboratively as a team are emerging as a key technology for future missions with high degrees of uncertainty such as extraterrestrial missions and terrestrial exploration in unmapped environments.

\begin{figure}[!t]
  \centering
  \includegraphics[width=1.0\linewidth]{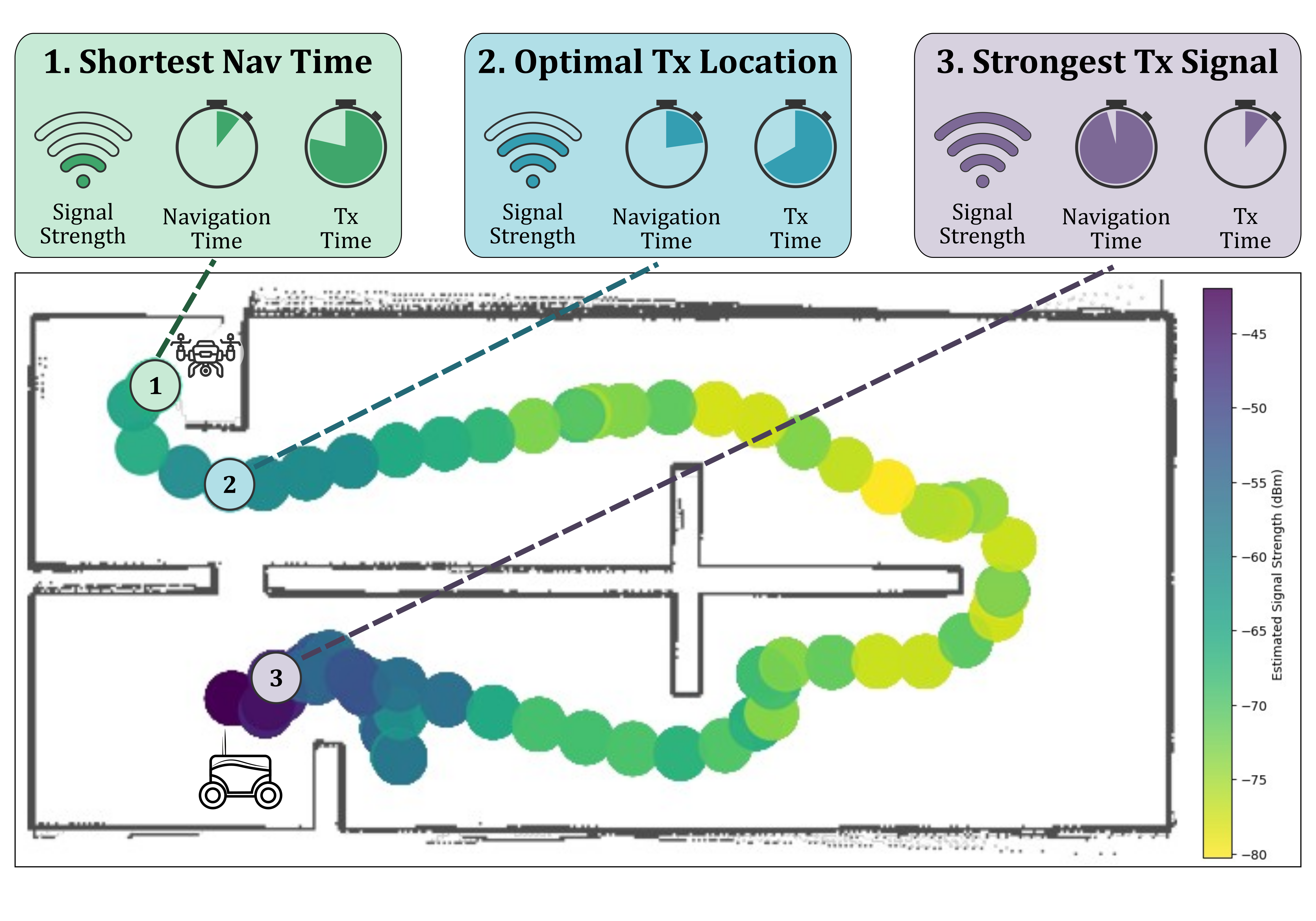}
  \caption{The Scout-Specialist robot team starts co-located in a simulated cave environment. The Scout flies ahead to scan the area, taking signal strength estimates to assess the communication link strength between the two robots. Shown here is the top down view of the signal strength estimations generated by the Scout vehicle as it explores the environment. A small opening provides an opportunity for Scout to transmit to the Specialist with higher bandwidth signal, even though it cannot traverse through the opening. From the samples the Scout has taken, a communication-aware motion planning algorithm determines the best trade-off between signal strength for a data payload size and the time cost of navigating to different locations before transmitting.}
  \label{fig:rf-on-map}
\end{figure}

By leveraging distributed intelligence, multi-agent systems can perform tasks such as terrain mapping and environmental analysis with greater efficiency than individual robots \cite{kim2014cooperative, Ropero2019}. Their ability to coordinate, self-organize, and adapt to unforeseen conditions makes them ideal for exploring scientifically rich targets including caves, tunnels, infrastructure, the Moon, Mars and beyond. Furthermore, multi-agent systems enhance mission resilience, as the failure of a single robot does not jeopardize the entire operation \cite{medos_agu, onair-aaai}.

When exploring and navigating through unmapped, cave-like environments, radio-frequency (RF) communication challenges are prevalent. This domain proved extremely challenging for autonomous robots during the DARPA Subterranean Challenge (SubT), a robotics competition inspiring new approaches for navigation, search, and artifact detection in complex underground environments \cite{ackerman2022robots-subt, biggie2023flexible, orekhov2022darpa-subt, ebadi2023slam-extreme-subt}. 
Inspired by the heterogeneous multi-agent systems deployed in the SubT challenge, we evaluate communication-aware exploration using models drawn from search-and-rescue robotics \cite{kitano1999robocup-sandr-robots, bogue2019disaster-sandr-robots, murphy2008search-and-rescue-robots} and prior paired-robot navigation approaches \cite{kim2014cooperative, Ropero2019}.

In particular, we focus on a ``Scout-Specialist'' paradigm of multi-agent systems as a representative example of multi-agent system that benefits from communication-aware planning. In this paradigm, both Scouts and Specialists are equipped for mapping, terrain estimation, and exploration, and there are generally one or more Scouts per Specialist. However, only the Specialists have the sensors necessary to complete the mission goal of scientific analysis. Consequently, the Scout agents are lightweight, low-cost, and agile, while the Specialists are less agile and must be more risk averse, traveling more cautiously. Specialists face significant traversal risk from incomplete terrain information, motivating Scouts to ``scout out'' potential mission targets, identify hazards, and relay high-fidelity updates. The Scout agent can mitigate the navigation risks to the Specialist by identifying areas of increased hazard and relaying this information to the Specialist, enhancing the Specialist’s traversability graph with hazard-aware updates which enables it to favor lower-risk paths.


\begin{figure}
    \centering
    \includegraphics[width=1\linewidth]{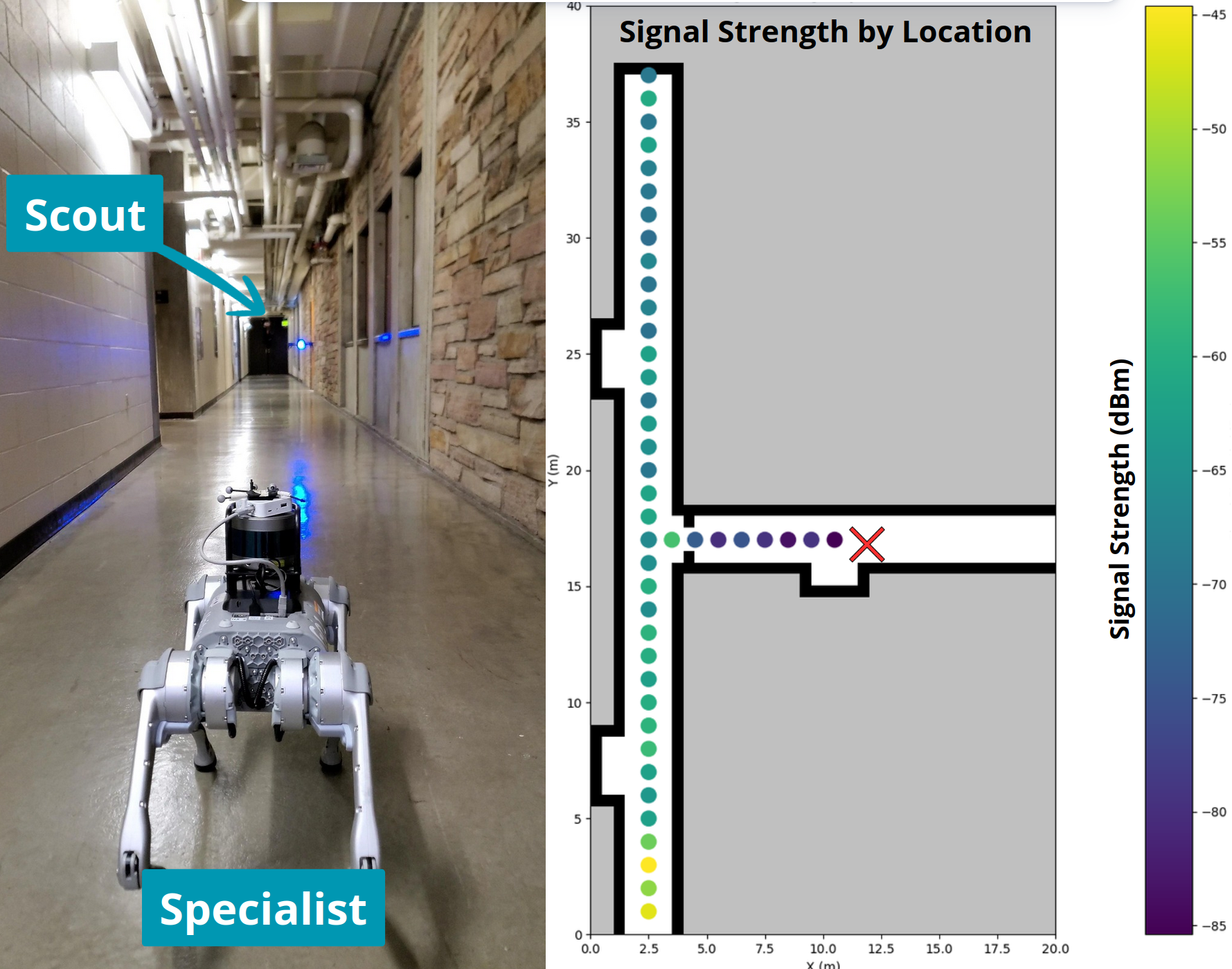}
    \caption{We collect signal strength measurements between two hardware platforms in a narrow corridor for baseline signal strength ranges in a real-world deployment. In line of sight settings, the signal strength remains strong for 40 meters. However, as soon as the Scout agent turns a corner and enters a perpendicular hallway, the signal strength rapidly declines, eventually resulting in a lost connection (depicted with a red x). The observed signal drop-off highlights the importance of adaptive transmission strategies that account for payload size and connectivity when deploying coordinated multi-agent systems.}
    \label{fig:drone-go1-hallway-signal-strength-map}
\end{figure}

Not all areas of an environment pose the same challenge for navigation. High-risk regions, such as complex terrain or cluttered passages, require detailed representations for a Specialist to build a reliable traversability graph, whereas low-risk regions, like flat ground, can be captured at lower resolution. Correspondingly, the amount of data needed to convey this information varies: complex areas demand high-resolution maps or short video clips for 3D reconstruction, while simpler regions can be represented with small, coarse updates. Transmitting larger data payloads efficiently requires closer proximity or stronger signal quality, whereas smaller updates can be shared over longer distances or weaker signal strengths.

This work leverages these trade-offs by modeling adaptive, communication-aware exploration. Scouts select transmission points based on predicted signal quality and the size of the information to be shared, balancing communication reliability against the cost of backtracking to ensure timely delivery of critical updates. Following Ropero et al.~\cite{Ropero2019}, we implement a Scout-Specialist team using a UAV as the Scout and a UGV as the Specialist. The UAV provides a bird’s-eye perspective, capturing high-fidelity information about complex regions that the ground vehicle cannot observe directly. In our experiments, this team cooperatively explores three distinct subterranean environments representing common cave geometries.

We introduce a novel radio communication-aware motion planning algorithm that drives efficient exploration for heterogeneous robot teams. We evaluate its performance in simulated subterranean environments, demonstrating shorter traversal distances and improved exploration efficiency compared to standard multi-robot communication strategies. Our method is grounded in signal strength measurements collected from real-world hardware experiments Figure~\ref{fig:drone-go1-hallway-signal-strength-map}.

\section{Related Works}

Early multi-robot exploration research focuses on improving coverage efficiency and coordination without considering resource constraints. Multi-robot teams independently explore different sections of an environment using frontier-based exploration and then reconvene at a rendezvous point to share their maps \cite{burgard2003collaborative, Burgard2005, Yamauchi1998}. This decentralized approach enables teams to map unknown spaces faster. Later work weighs exploration benefit against travel cost to effectively distribute robots for minimizing total mission time \cite{Burgard2005}. 

\textbf{Communication Constraints.} As multi-agent teams venture into communication limited environments, bandwidth constraints and efficient data-sharing strategies become more important. Anderson et al. \cite{Andersone2013} reduce the occupancy-grid resolution to curb the wireless traffic, enabling more frequent map updates between several robots. In PropEM-L \cite{Clark2022}, robots use online radio-signal modeling to plan routes that preserve connectivity in unstructured environments. Complementary approaches such as ACHORD leverage deployable relays and stream prioritization to sustain critical data exchange despite intermittent links \cite{saboia2022achord}. This work extends these strategies by demonstrating how signal strength informed motion planning can make heterogeneous robot teams more efficient in exploration tasks.

\textbf{Energy Constraints.} Prior work has explored energy-aware coordination for multi-robot teams. Ropero et al. \cite{Ropero2019} demonstrate a UGV–UAV team where the ground robot recharges the quadcopter to extend flight endurance. Maresca et al. \cite{Maresca2025} introduce REACT, dynamically reallocating search regions to minimize redundant travel and preserve battery, while Munir et al. \cite{Munir2024} scale robot coverage based on remaining energy to maximize exploration before recharging. Seewald et al. \cite{Seewald2024} schedule staggered recharging to maintain continuous coverage. Building on this literature, our approach leverages signal strength to select energy-efficient locations for data transmission between robots.

\textbf{Multi-Robot Coordination Strategies.} Planning-based coordination has been studied extensively for exploration under constrained communication. Frameworks that model teammates’ internal states allow decentralized task execution without continuous connectivity \cite{bramblett2023epistemic}, and even limited-range, infrequent communication can improve coverage compared to fully disconnected teams \cite{kedege2022multi}. Surveys highlight the need for adaptable, event-triggered communication policies that handle dynamic topology, bandwidth limits, and heterogeneous agent roles \cite{amigoni2017multirobot}. Some works focus on rendezvous strategies, bringing agents back together to exchange information \cite{wurm2010coordinated-marsupial-teams, de2010selection-rendezvous-points}, which can force unnecessary backtracking. In contrast, our communication-aware motion planning algorithm dynamically selects transmission locations based on predicted signal strength, balancing backtracking distance and data exchange to enable more efficient multi-robot exploration.


\section{Methods}

\subsection{Method Overview}
We investigate the benefits of communication-aware motion planning through the use of multi-robot, Scout-Specialist exploration scenarios. The core of our contributed approach is an adaptive, communication-aware motion planning algorithm which allows Scouts to balance exploration efficiency with data transmission costs when synchronizing with Specialists. Each environment in our evaluation contains a location that triggers a data-sharing event (e.g., a scientific point of interest or terrain hazard), indicating that the Scout must transmit to the Specialist. To select the appropriate transmission point, the Scout combines signal strength estimation with a task disruption score that captures environmental factors like transit time to the communication point. While this work utilizes an established signal model, outlined in Section~\ref{sec:signal-strength-modeling}, the contributed method is agnostic to signal propagation model and communication protocol so long as the duration of data transmission can be estimated given a measured signal strength. The proposed adaptive strategy allows a Scout to minimize backtracking while ensuring that the Specialist receives transmission at the required fidelity. 

\subsection{Software System Overview}\label{method-overview}
\begin{figure}
    \centering
    \includegraphics[width=1.0\linewidth]{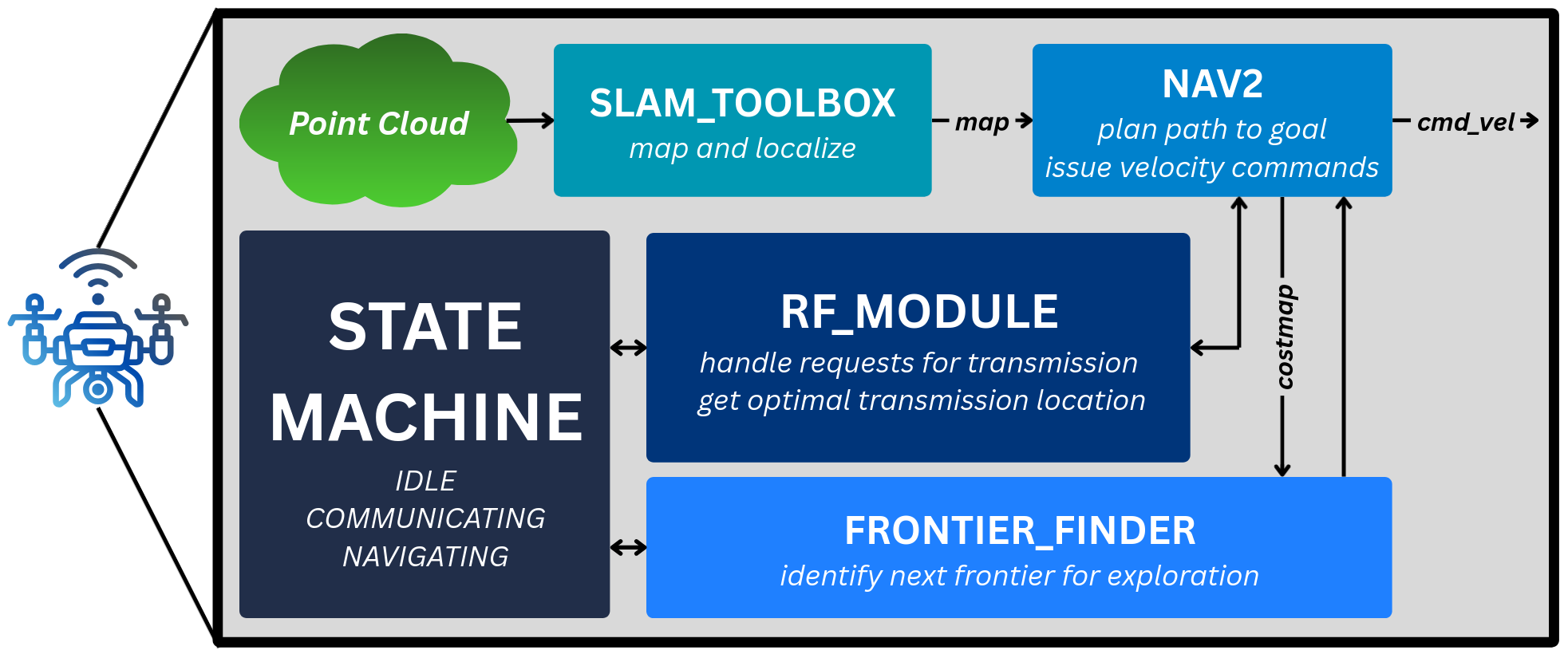}
    \caption{The system uses a custom state machine, frontier finder, and adaptive-RF module to evaluate the communication-aware motion planning algorithms described in Section~\ref{sec:evaluation}. The custom software components integrate with standard ROS2 packages: NAV2 and slam\_toolbox. Each robot runs the same software stack and communication takes place over ROS2 message exchange when the agents are within transmission range of one another.}
    \label{fig:software-stack}
\end{figure}

Each platform runs the same software stack shown in Figure \ref{fig:software-stack}. The system consists of modules for frontier finding, adaptive RF communication, and coordination. These components integrate with ROS2’s NAV2 \cite{vslamComparison2021nav2, macenski2020marathon2nav2, macenski2024smacNAV2, macenski2023surveyNAV2} and SLAM Toolbox \cite{macenski2021slamtoolbox} packages which perform path planning and simultaneous localization and mapping (SLAM) respectively. A simple state machine runs on board each agent to manage tasking. All software is available at \textit{insert repo URL here}.


\subsection{Frontier Finding} \label{sec:frontier-finding}
We consider operations in environments that have not been mapped apriori for evaluating ART, where a Scout needs to make a determination about where to navigate. A wavefront-based frontier detection strategy identifies candidate exploration targets from a 2D occupancy grid map \cite{keidar2012robotfrontierfinding, keidar2014efficientfrontierfinding}. Frontier candidates are filtered by their distance from the robot, discarding those that are too close (within 0.5 m) and those beyond the current search radius. If no valid candidates are found, the search radius is expanded and the process repeated. Finally, the resulting frontier candidates are sorted by distance, and up to five of the nearest are selected as exploration goals. If no frontier candidates are found with an expanded search radius, the search expands to the global costmap. When no frontiers remain, exploration is complete.

\subsection{Signal Strength Modeling} \label{sec:signal-strength-modeling}
Communication-aware motion planning requires an estimate of the signal strength between agents. A simple approach to signal strength estimation is to apply the Friis Transmission equation which quantifies power at the receiver in free space conditions \cite{Johnson1993}:

{\footnotesize
\begin{equation}
P_r(d)=P_t G_t G_r \left(\frac{\lambda}{4\pi d}\right)^2,
\label{eq:friis}
\end{equation}
}
where $P_t$ is the transmit power, $G_t$ and $G_r$ are the transmitter and receiver antenna gains, $\lambda$ is the carrier wavelength, and $d$ is the separation distance. However, this free-space model does not capture the attenuation introduced by obstacles such as walls, rubble, or geological features which interfere with signal transmission. Therefore, a more realistic representation known as the Log-Distance Path-Loss model \cite{Rappaport2002, Goldsmith2005} is used. This model accounts for obstacles and multi-path attenuation which grows logarithmically with distance:

{\footnotesize
\begin{equation}
PL(d)=PL(d_0)+10\,n\log_{10}\!\left(\frac{d}{d_0}\right)+X_\sigma,
\label{eq:logdistance}
\end{equation}
}

where $PL(d)$ is the path loss (dB) at distance $d$, $PL(d_0)$ is the reference loss at a close‑in distance $d_0$, $n$ is the environment‑dependent path‑loss exponent, and $X_\sigma$ is a zero‑mean Gaussian random variable (dB) representing shadow fading with standard deviation $\sigma$. Following El-Khalad et al. \cite{El-Khalad2022logdistancepathloss}, a value of n=3 is selected to approximate attenuation in cluttered indoor environments, the closest approximation for enclosed, cave-like environments.

To further improve model fidelity in environments with complex geometry, such as caves, the propagation path length is estimated as the signal path rather than a simple Euclidean distance. This accounts for indirect transmission paths created by features such as archways or skylights, which may permit communication even when physical traversal is infeasible. The signal path distance is computed using the A* algorithm \cite{hart1968A-star-algo}, enabling the model to capture transmission windows that occur despite navigational constraints (e.g., as seen in the Window environment in Figure~\ref{fig:cave-inspiration}).

\subsection{Disruption Score} \label{sec:disruption-score}
A disruption score, $d\_score$, is used to quantify the cost associated with interrupting exploration for data transmission. The score captures the time required for the Scout agent to travel to a candidate transmission point, transmit a data payload of a given size, and return to the previous exploration frontier. It is expressed as:

\begin{equation}
\text{$d\_score$} = 
t_{\text{to\_location}} +
t_{\text{transmitting}} +
t_{\text{return\_to\_frontier}}
\end{equation}

Lower disruption scores correspond to transmission locations that balance communication quality with impact on exploration progress. This work utilizes common WiFi standards for transmission time estimation as outlined in Section~\ref{sec:modeling-wifi-trans-time-signal-strength} but the method is agnostic to the communication protocol so long a transmission duration can be obtained.

\subsection{Data Payload Size Estimation} \label{sec:data-payload-sizes}
Data size requirements for different communication events are categorized into four levels, summarized in Table~\ref{table:data-size-estimation}. These estimates are derived from common robotic sensor outputs and data formats.

At the lowest level, only lightweight telemetry such as status messages are required. A single 3D coordinate expressed as a ROS2 geometry\_msgs/msg/Point message occupies on the order of a few dozen bytes \cite{ros_geometry_msgs_point}; a conservative estimate of 1~kB is used to account for message overhead.

Low-bandwidth visual updates correspond to compressed imagery, with a typical 640$\times$480 JPEG image requiring approximately 100~kB. Higher-fidelity perception data, such as 1280$\times$720 images or LiDAR sweeps, typically occupy several megabytes and are conservatively estimated at 10~MB \cite{IntelRealSenseD455, geiger2013visionkittidataset}. Large-scale data transfers, including multi-frame lidar scans or short video segments, can exceed 100~MB and are modeled at this size \cite{geiger2013visionkittidataset}.

\begin{table}[ht]
  \centering
  \caption{Qualitative size estimates for common robotics data}
  \label{table:data-size-estimation}
  \scriptsize
  \begin{tabular}{@{}cccc@{}}
    \toprule
    Tx Level & Representation & Resolution / Description & Approx. Size \\ \midrule
    0 & Telemetry/Status & ROS2 Status Msg & 1 kB \\
    1 & Compressed Image & 640\,$\times$\,480 JPEG & 100 kB \\
    2 & High-Res Image/LiDAR & 1280\,$\times$\,720 JPEG & 10 MB \\
    3 & Video Stream & Multi-frame LiDAR & 100 MB \\
    \bottomrule
  \end{tabular}
\end{table}

\subsection{Modeling Transmission Time Based on Signal Strength \label{sec:modeling-wifi-trans-time-signal-strength}}

Transmission time between agents over a radio-frequency link is modeled as a function of payload size and received signal strength indicator (RSSI). The transmission time is given by:

{\footnotesize
\begin{equation}
t = \frac{S}{C}
\end{equation}
}

where $t$ is the transmission time in seconds, $S$ is the data payload size in bits, $C$ is the channel capacity in bits per second (bps).

Channel capacity depends on the signal-to-noise ratio (SNR). Using the Shannon theorem \cite{Rappaport2002}:

{\footnotesize
\begin{equation}
C = B \cdot \log_2(1 + \text{SNR})
\end{equation}
}

where: $B$ is the channel bandwidth in Hz, $\text{SNR}$ is the linear signal-to-noise ratio, defined as \cite{Rappaport2002}:

{\footnotesize
\begin{equation}
    \text{SNR} = \frac{P_\text{signal}}{P_\text{noise}}
\end{equation}
}

Here, $P_\text{signal}$ is the received signal power in milliwatts, and $P_\text{noise}$ is the noise floor power in milliwatts.

The received signal power, $P_\text{signal}$, and noise floor power, $P_\text{noise}$, are computed from RSSI and noise floor measurements (in dBm) using:

{\footnotesize
\begin{equation}
P_\text{signal}~[\text{mW}] = 10^{\frac{\text{RSSI [dBm]}}{10}}
\end{equation}
}

{\footnotesize
\begin{equation}
P_\text{noise}~[\text{mW}] = 10^{\frac{\text{NoiseFloor [dBm]}}{10}}
\end{equation}
}

A typical noise floor used for data transmission is approximately $-88$ dBm \cite{lopez2021performance-noisefloor}. Signals near or below this level result in unreliable communication. The communication model combines several components to capture the interaction between environment, signal propagation, and transmission performance. The Log-Distance Path-Loss model accounts for attenuation due to obstacles, while signal path estimation through A* incorporates geometric complexity. Payload size categories reflect realistic robotic data requirements, and transmission time is modeled as a function of received signal strength, linking environmental conditions directly to communication performance. Combining these elements enables estimation of data transfer duration under the constraints of subterranean exploration scenarios.

\subsection{Algorithm Descriptions}

The following sections contain detailed descriptions of the high-level mission algorithm, the Adaptive-RF Transmission algorithm (ART), and a high-fidelity data transmission variant ART-SST.

\subsubsection{High-Level Mission}
Algorithm \ref{alg:high-level-algo} shows the higher level algorithm for the mission. The Scout and Specialists begin exploration with the Scout moving out ahead of the Specialist using the frontier finding strategy described in Section \ref{sec:frontier-finding}.

\begin{algorithm}[h]
\footnotesize
\DontPrintSemicolon
\caption{High‑level Collaborative Exploration}
\label{alg:high-level-algo}
\KwIn{Maximum mission time $T$}
\vspace{0.5\baselineskip}
$current\_t \leftarrow 0$

\While{$current\_t < T$}{
    \tcp{Scout explores frontier regions}
    \texttt{ScoutExploreDetectTransmit}()
    \BlankLine
    \tcp{Specialist rover awaits tasking}
    \If{\texttt{NewGoalReceived}($g$)}{
        \texttt{NavigateToGoal}($g$)
        \texttt{TransmitStatusToScout}()

        \uIf{\texttt{ReachedGoal}()}{
            \texttt{WaitAtGoalPoint}()
        }\Else{
            \texttt{HandleFailureCase}()
        }
    }
}
\end{algorithm}

\subsubsection{Signal Strength Guided Collaborative Communication}
The Scout employs Algorithm \ref{alg:scout-algo-rssi} during exploration. While navigating to new frontiers and exploring the space, the Scout detects locations within the environment that trigger a transmission to the Specialist. Each transmission receives an importance level which is directly associated with a corresponding data payload size, reflecting the amount of information necessary for the Specialist to make informed decisions. Higher-importance events require larger payloads to preserve necessary detail. Scout classifies the importance level on a scale of 0-3 with 0 requiring the lowest data payload transmission (lowest importance) and 3 requiring the largest payload transmission (highest importance).

Because transmission time depends on both payload size and link quality, the Scout evaluates whether the signal strength at its current position is sufficient for the required payload. If the signal is too weak, the Scout searches its recorded signal strength map to identify candidate transmission locations that satisfy the required bandwidth threshold. Among these, it selects the location with the lowest disruption score (as defined in Section~\ref{sec:disruption-score}), balancing transmission quality against the cost of backtracking. The Scout then navigates to that location to complete the transmission before resuming exploration.

\begin{algorithm}[h]
\footnotesize
\DontPrintSemicolon
\caption{Exploration with Signal Strength Guided Collaborative Transmission}
\label{alg:scout-algo-rssi}
\KwIn{Maximum mission time $T$}
\KwOut{Environment map $M$}
\vspace{0.5\baselineskip}

$current\_t \leftarrow 0$\;

\While{$current\_t < T$}{
    $\mathcal{F} \leftarrow$ \texttt{IdentifyFrontiers}()\;
    \If{$\mathcal{F} = \emptyset$}{
        \tcp{No frontiers remaining}
        \textbf{break}\;
    }
    $g^\star \leftarrow$ \texttt{SelectFrontier}($\mathcal{F}$)\;
    $rssi \leftarrow$ \texttt{SampleSignalStrength}()\;
    $\mathcal{S} \leftarrow$\texttt{LogSignalStrength}($rssi, \texttt{CurrentPose}()$)\;

    \While{\texttt{CurrentPose}() $\neq g^\star$ }{
        $M \leftarrow$ \texttt{UpdateMap}()\;
        $\phi \leftarrow$ \texttt{DetectFeature}()\;

        \If{\texttt{RequiresTransmission}($\phi$)}{
            $h \leftarrow$ \texttt{ComputeDataSize}($\phi$)\;
            \If{\textbf{not} \texttt{InTxRange}($h$)}{
                $scout_{\text{pose}} \leftarrow$ \texttt{CurrentPose}()\;
                $specialist_{\text{pose}} \leftarrow$ \texttt{GetSpecialistLocation}()\;

                 \tcp{ See Algorithm \ref{alg:get-best-tx-location}}
                $best\_tx\_loc \leftarrow$ \texttt{GetBestTXLocation}()\;
                \texttt{NavigateTo}($best\_tx\_loc$)\;
            }

            \texttt{TransmitPayload}($h$)\;
            \texttt{NavigateTo}($scout_{\text{pose}}$)\;
        }

        \texttt{ExploreWithGoal}($g^\star$)\;
        $current\_t \leftarrow$ \texttt{Now}()\;
    }
}

\tcp{Exploration Complete or Mission Time Exhausted}
\Return $M$\;
\end{algorithm}

\subsubsection{Computing the Best Transmission Location}
Using the Scout's current position and the last known location of the Specialist, the Scout uses Algorithm \ref{alg:get-best-tx-location} to find the best location for transmitting to the Specialist. Each transmission category (Table \ref{table:data-size-estimation}) has a corresponding threshold which the signal strength must be above in order to successfully transmit the data consistent with common WiFi signal strengths \cite{oscium2025wifi}. ART uses a minimum signal threshold for a usable WiFi network while ART-SST enforces strict signal strength thresholds by data payload size.

The threshold is applied to all Scout's signal strength estimates, $\mathcal{S}$. The set $\mathcal{S}$ is composed of tuples of signal strength and the location at which they were sampled, $(s_i, p_i)$, over map, $\mathcal{M}$. Only the candidates above the threshold are kept. For each candidate, the disruption score is computed as described in Section \ref{sec:disruption-score}. The candidate point with the lowest disruption score is chosen because it represents the required signal strength point that is least costly in terms of navigation transit time and transmission delivery time.

The scout then navigates to the transmission location and makes the transmission. Upon receiving the acknowledgment of a received transmission from the Specialist, Scout returns to its previously saved pose to restart exploration. This continues until there are no more frontiers and the environment is considered fully explored.

\begin{algorithm}[h]
\footnotesize
\DontPrintSemicolon
\caption{GetBestTXLocation}
\label{alg:get-best-tx-location}
\KwIn{Signal strength estimates $\mathcal{S} = \{(s_i, p_i)\}$; data payload size $h$; environment map $M$}
\KwOut{Best transmission location $(s^\star, p^\star)$}
\vspace{0.5\baselineskip}

$\tau \leftarrow$ \texttt{GetSignalThreshold}($h$)\;
\tcp{Get candidates above the threshold}
$\mathcal{C} \leftarrow \{(s_i, p_i) \in \mathcal{S} \mid s_i \geq \tau \}$

\ForEach{$(s, p) \in \mathcal{C}$}{
    $d(p) \leftarrow$ \texttt{ComputeDisruptionScore}($p, M$)\;
}
\tcp{Select least disruptive as the best}
$(s^\star, p^\star) \leftarrow \arg\min_{(s,p) \in \mathcal{C}} d(p)$\;

\Return $(s^\star, p^\star)$\;
\end{algorithm}

\subsection{Simulation Specifications}\label{sim-specs-reqs}
The simulator is Nvidia IsaacSim v4.5 running on a host machine with an RTX4090 GPU. Two different robot platforms serve as Scout and Specialist: an Iris UAV and a NOVA Carter ground platform. These platforms are chosen for their differing capabilities, but the algorithmic approach is platform agnostic and can be applied to any robot teams able to measure communication link strength.

\section{Evaluation} \label{sec:evaluation}
The proposed approach is evaluated against three comparative methods to assess its effectiveness in communication-constrained exploration scenarios. The four algorithms examined are:

\begin{itemize}
    \item Adaptive-RF Transmission (ART): The adaptive algorithm without enforced minimum thresholds, selecting transmission locations based solely on minimizing the disruption score.
    \item Adaptive-RF Transmission – Signal Strength Threshold (ART-SST): The adaptive algorithm with enforced minimum signal thresholds per payload size.
    \item Minimum Signal Strength Condition (MSSC): A naive strategy where the Scout transmits from the first location where a minimum signal strength threshold is detected.
    \item Full Rehome Condition (FRC): A conventional rendezvous approach in which the Scout returns to within 1~meter of the Specialist for every data transmission.
\end{itemize}

Each algorithm is evaluated under identical conditions across three environments representative of subterranean exploration scenarios. Each evaluation episode begins with the robot team co-located at a common starting position within the environment. The Scout advances ahead of the Specialist to explore unknown regions, incrementally constructing a map, sampling signal strength estimates, and transmitting data according to the conditions defined by the algorithm under test. An episode terminates when two criteria are satisfied: (1) all identified frontier regions have been explored, and (2) all pending data transmissions have been completed. This termination condition ensures that both exploration and communication objectives are fully evaluated within a single episode.

\subsection{Environments} \label{sec:environments}
\begin{figure}
    \centering
    \includegraphics[width=1.0\columnwidth]{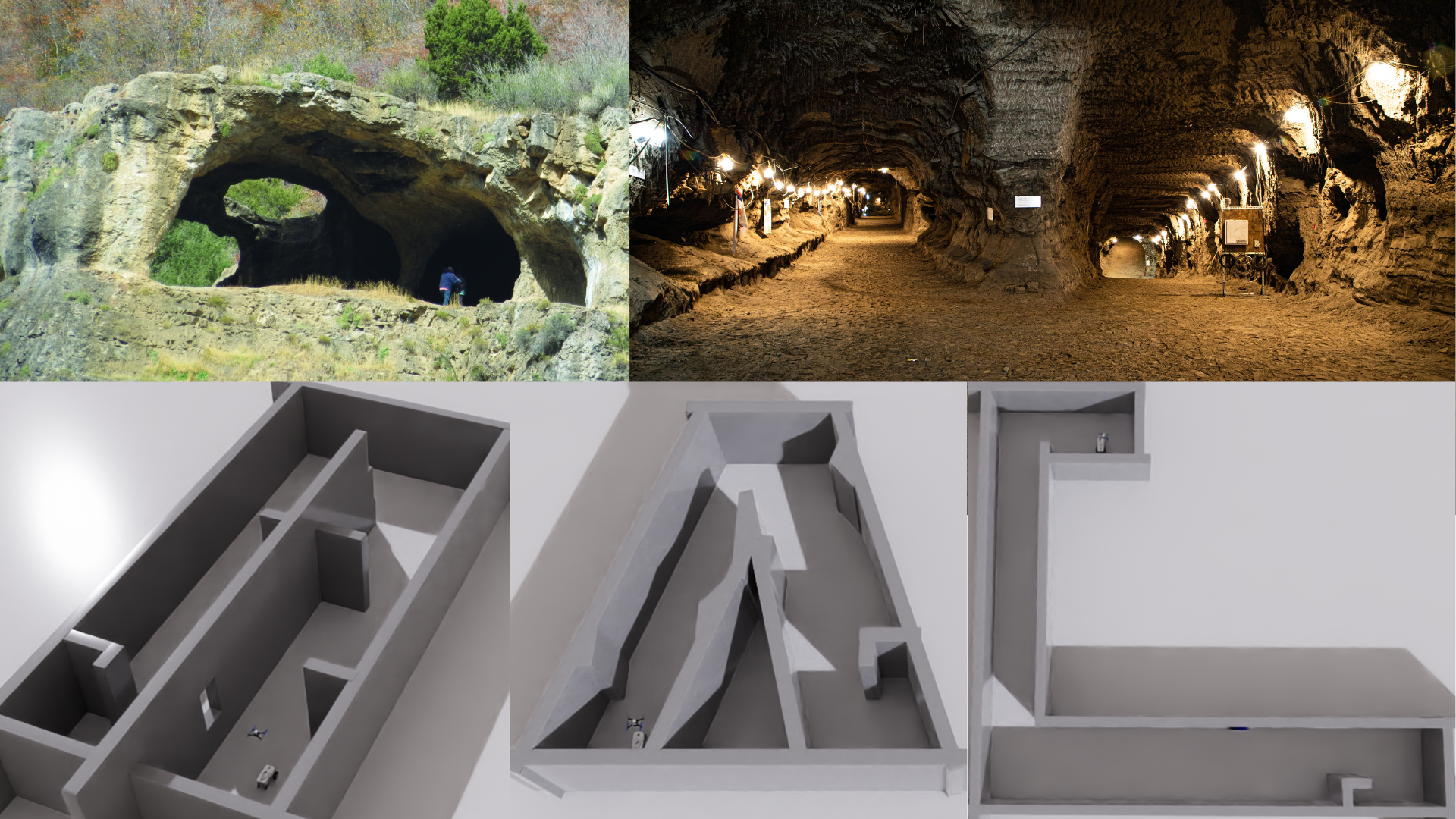}
    \caption{Top left: visible skylight at Wind Cave in Logan, Utah, USA \cite{Carney2020_WindCaveGeoSights}. Top right: Permafrost Research Center in Fairbanks, Alaska, USA \cite{USACE2025_PermafrostTunnelPhoto}. These real-world environments inspire the design of simulated testing domains used in this work. The window environment (bottom left) models the skylight geometry while the Y-junction environment (bottom middle) reflects the branching structure of the permafrost tunnel. A third environment (bottom right), a long and narrow tunnel, forces agents out of communication range, enabling evaluation of algorithm performance under communication-denied conditions.}
    \label{fig:cave-inspiration}
\end{figure}

Three environments are constructed to reflect cave geometries observed in natural and human-made subterranean spaces \cite{Carney2020_WindCaveGeoSights, USACE2025_PermafrostTunnelPhoto}. The first is a window environment, inspired by cave archways and skylights, which contains non-traversable openings that permit signal propagation between agents without providing a navigable path. The second is typical Y-junction often seen in mining tunnels and natural cave environments which requires exploration decisions at intersections. The third environment is a standard tunnel with a separate entrance and exit which ensures that the two agents will be out of communication range for portions of the exploration task. Real cave geometries which inspired these choices and the simulation environments used for testing can be viewed in Figure~\ref{fig:cave-inspiration}.

\subsection{Adaptive-RF Transmission (ART)}
This condition utilizes Algorithm \ref{alg:scout-algo-rssi} to adaptively determine how close to the Specialist that the Scout should get to make the transmission by choosing a transmission location that balances transmission duration with backtracking navigation time. The Scout navigates to within that range, shares the update, and then returns to the previously saved navigation frontier to continue exploration.

\subsection{Adaptive-RF Transmission with Signal Strength Thresholds (ART-SST)}
The ART-SST condition extends the adaptive transmission approach by introducing explicit signal strength requirements for each payload category. In this setting, Algorithm~\ref{alg:scout-algo-rssi} operates under the constraint that a transmission can only occur at locations where the received signal strength exceeds a payload-specific minimum threshold. These thresholds model operational requirements in scenarios where data fidelity or reliability must be guaranteed, such as the transmission of high-resolution perception data. The Scout identifies the transmission point that satisfies the signal requirement while minimizing disruption cost, transmits the payload, and resumes exploration. ART-SST enables analysis of how strict communication requirements affect overall exploration efficiency.

\subsection{Minimum Signal Strength Condition (MSSC)}
The MSSC condition serves as a simple baseline representing heuristic communication strategies commonly used in multi-robot exploration. Under this condition, the Scout initiates transmission as soon as a signal above a minimal usable threshold is detected, without considering transmission time or backtracking cost. This approach approximates a simple “first viable signal” policy and serves as a practical lower bound for performance. Following transmission, the Scout returns to its previous frontier and continues exploration. MSSC allows evaluation of how much performance gain is achieved by incorporating simple communication-aware heuristics into motion planning.

\subsection{Full Rehome Condition (FRC)}
The Full Rehome Condition (FRC) represents a classical rendezvous-based coordination strategy. In this baseline, the Scout navigates back to within 1~meter of the Specialist for each transmission event before resuming exploration. This approach, widely used in early multi-robot exploration work \cite{wurm2010coordinated-marsupial-teams, de2010selection-rendezvous-points}, prioritizes communication reliability over efficiency but often incurs significant backtracking costs. As such, it serves as a useful reference for assessing the efficiency gains provided by adaptive communication strategies such as ART and ART-SST.

\section{Results \& Discussion} \label{sec:results-discussion}

\begin{table*}[]
  \centering
  \resizebox{\textwidth}{!}{%
  \begin{tabular}{@{}c|c|cc|cc|cc@{}}
    \toprule
    & & \multicolumn{2}{c|}{\textbf{Tunnel}} & \multicolumn{2}{c|}{\textbf{Window}} & \multicolumn{2}{c}{\textbf{Y-Junction}} \\ 
    \cmidrule(lr){3-4} \cmidrule(lr){5-6} \cmidrule(lr){7-8}
    \textbf{TX} & \textbf{Algorithm} & Path Dist. & Time & Path Dist. & Time & Path Dist. & Time \\
    \midrule

    \multirow{4}{*}{0} 
      & ART       & \textbf{45.294 $\pm$ 2.827} & \textbf{96.502 $\pm$ 7.786} & \textbf{44.252 $\pm$ 2.710} & \textbf{96.262 $\pm$ 10.603} & \textbf{33.000 $\pm$ 4.306} & \textbf{82.702 $\pm$ 12.783} \\
      & ART-SST   & 45.387 $\pm$ 3.780 & 97.080 $\pm$ 5.608 & 46.396 $\pm$ 7.053 & 112.017 $\pm$ 18.103 & 34.288 $\pm$ 1.959 & 85.283 $\pm$ 6.024 \\
      & MSSC      & 49.902 $\pm$ 3.553 & 103.567 $\pm$ 7.409 & 47.564 $\pm$ 6.098 & 108.313 $\pm$ 10.652 & 33.979 $\pm$ 4.083 & 83.677 $\pm$ 13.722 \\
      & FRC       & 99.267 $\pm$ 2.264 & 180.617 $\pm$ 6.951 & 105.301 $\pm$ 3.135 & 195.807 $\pm$ 16.841 & 59.430 $\pm$ 2.121 & 120.653 $\pm$ 6.725 \\
    \midrule

    \multirow{4}{*}{1} 
      & ART       & \textbf{47.925 $\pm$ 2.653} & \textbf{100.198 $\pm$ 6.855} & \textbf{45.181 $\pm$ 2.921} & \textbf{98.513 $\pm$ 6.541} & \textbf{32.864 $\pm$ 3.481} & \textbf{81.757 $\pm$ 9.800} \\
      & ART-SST   & 58.093 $\pm$ 3.737 & 116.020 $\pm$ 7.045 & 47.361 $\pm$ 5.539 & 111.765 $\pm$ 12.183 & 33.255 $\pm$ 1.593 & 86.062 $\pm$ 7.661 \\
      & MSSC      & 51.784 $\pm$ 3.739 & 108.737 $\pm$ 9.904 & 46.395 $\pm$ 3.215 & 110.087 $\pm$ 14.647 & 34.322 $\pm$ 4.183 & 90.813 $\pm$ 9.085 \\
      & FRC       & 98.734 $\pm$ 1.261 & 181.248 $\pm$ 4.554 & 108.243 $\pm$ 5.849 & 196.657 $\pm$ 22.328 & 58.306 $\pm$ 2.639 & 121.535 $\pm$ 8.488 \\
    \midrule

    \multirow{4}{*}{2} 
      & ART       & \textbf{45.513 $\pm$ 2.736} & \textbf{97.217 $\pm$ 7.075} & 47.171 $\pm$ 7.512 & \textbf{96.670 $\pm$ 13.878} & \textbf{33.244 $\pm$ 1.388} & \textbf{80.437 $\pm$ 7.927} \\
      & ART-SST   & 82.345 $\pm$ 2.132 & 153.853 $\pm$ 4.603 & \textbf{43.475 $\pm$ 1.337} & 102.175 $\pm$ 12.238 & 41.482 $\pm$ 7.247 & 97.243 $\pm$ 10.906 \\
      & MSSC      & 49.666 $\pm$ 2.716 & 108.172 $\pm$ 7.750 & 45.300 $\pm$ 4.095 & 104.450 $\pm$ 9.466 & 36.078 $\pm$ 3.343 & 93.438 $\pm$ 16.409 \\
      & FRC       & 98.937 $\pm$ 1.713 & 179.008 $\pm$ 4.268 & 106.842 $\pm$ 4.255 & 200.378 $\pm$ 7.761 & 58.638 $\pm$ 2.045 & 118.310 $\pm$ 3.419 \\
    \midrule

    \multirow{4}{*}{3} 
      & ART       & \textbf{48.005 $\pm$ 2.605} & \textbf{116.207 $\pm$ 8.540} & \textbf{47.854 $\pm$ 5.105} & \textbf{104.783 $\pm$ 14.846} & \textbf{34.684 $\pm$ 2.907} & \textbf{92.312 $\pm$ 13.455} \\
      & ART-SST   & 96.845 $\pm$ 1.382 & 180.007 $\pm$ 5.130 & 104.346 $\pm$ 2.862 & 193.650 $\pm$ 9.677 & 55.557 $\pm$ 3.432 & 120.228 $\pm$ 12.265 \\
      & MSSC      & 48.163 $\pm$ 2.844 & 118.927 $\pm$ 13.260 & 48.547 $\pm$ 4.952 & 115.852 $\pm$ 9.313 & 36.230 $\pm$ 3.488 & 94.732 $\pm$ 11.592 \\
      & FRC       & 98.372 $\pm$ 1.357 & 181.628 $\pm$ 5.718 & 108.040 $\pm$ 3.654 & 201.568 $\pm$ 15.756 & 61.564 $\pm$ 2.251 & 126.685 $\pm$ 7.353 \\
    \bottomrule
  \end{tabular}
  }
    \caption{Mean $\pm$ standard deviation of path distance (m) and exploration time (s) across all algorithms and payload sizes for three environments. Lower mean values indicate more efficient exploration. Bold values are the minimum mean in each column.}
  \label{table:combined-environments}
\end{table*}

All algorithms described in Section~\ref{sec:evaluation} are evaluated based on two primary performance metrics: percentage of the environment explored per unit of flight time and the total path distance traveled by the Scout agent. Experiments consist of 160 simulation runs in each environment, spanning the four algorithms defined in Section~\ref{sec:evaluation} and the four payload sizes outlined in Section~\ref{sec:data-payload-sizes}. Each configuration is executed in 10 independent trials.

\subsection{Overall Performance Trends}

Across all environments and payload sizes, the Adaptive-RF Transmission (ART) algorithm consistently achieves the lowest exploration time and shortest total path distance compared to the three baseline approaches (Table~\ref{table:combined-environments}). ART achieves up to 51.7\% faster exploration and reduces path distance by as much as 58.2\% relative to the Full Rehome Condition (FRC). These improvements result from ART’s dynamic balancing of signal quality and backtracking cost which enables selection of transmission points that minimize disruption to the primary exploration task.

\begin{figure}
    \centering
    \includegraphics[width=1.0\linewidth]{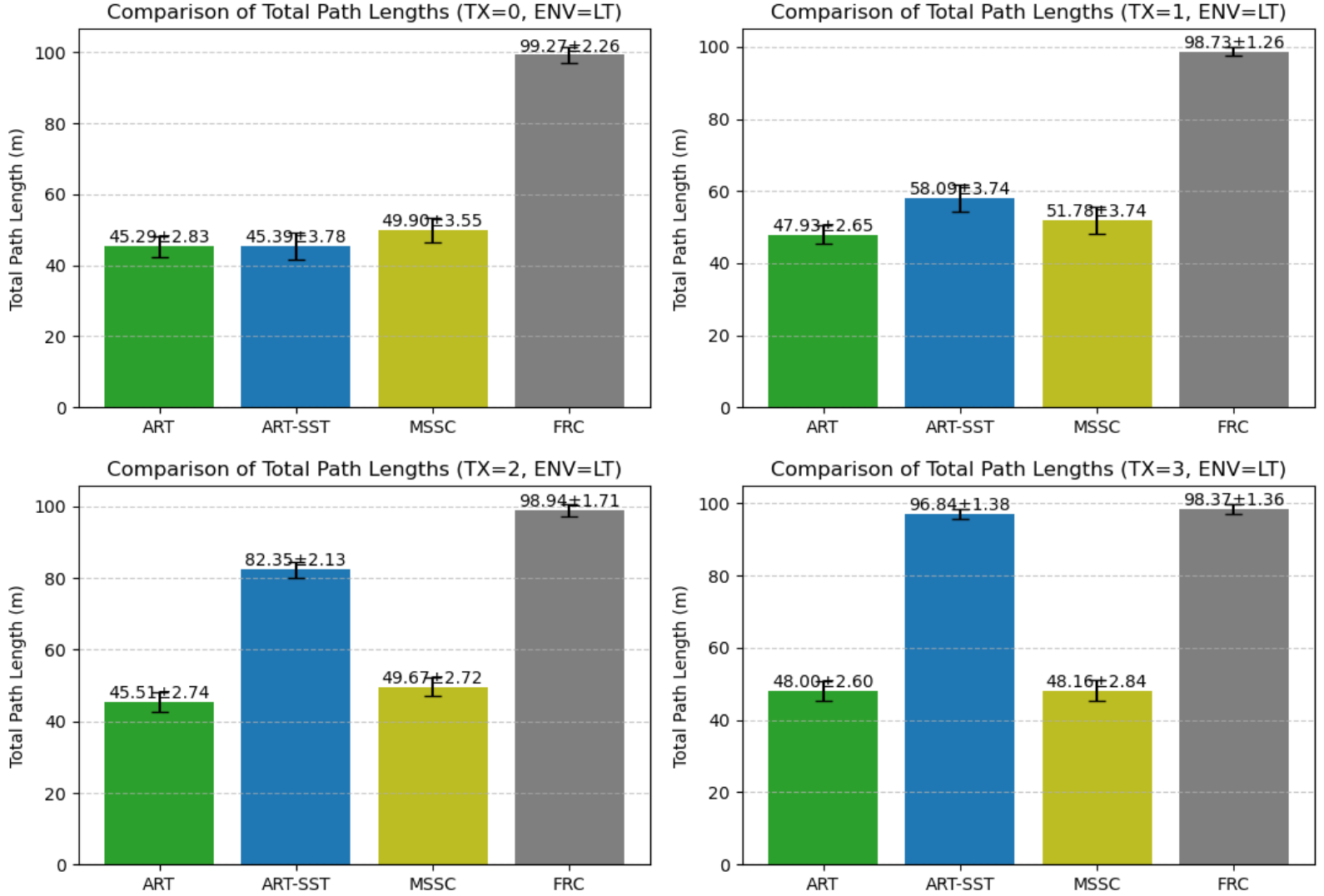}
    \caption{Comparison of mean total path lengths for all algorithms across all payload sizes (0–3) in the Long Tunnel (LT) environment. ART consistently achieves the shortest paths across all payload sizes, indicating more efficient exploration. ART-SST, MSSC, and FRC see substantially longer paths, with FRC performing the worst overall. Error bars represent one standard deviation.}
    \label{fig:lt-all-tx-path-dist}
\end{figure}

ART-SST introduces operational constraints through enforcement of minimum signal strength thresholds, which, for larger payload sizes, leads to longer backtracking distances and slower exploration relative to ART and MSSC. MSSC typically experiences extended transmission times due to weaker link quality. FRC, which requires full rendezvous for every transmission, produces the slowest exploration and longest paths, highlighting the inefficiency of fixed rendezvous strategies in communication-limited environments.

\begin{figure}
    \centering
    \includegraphics[width=1\linewidth]{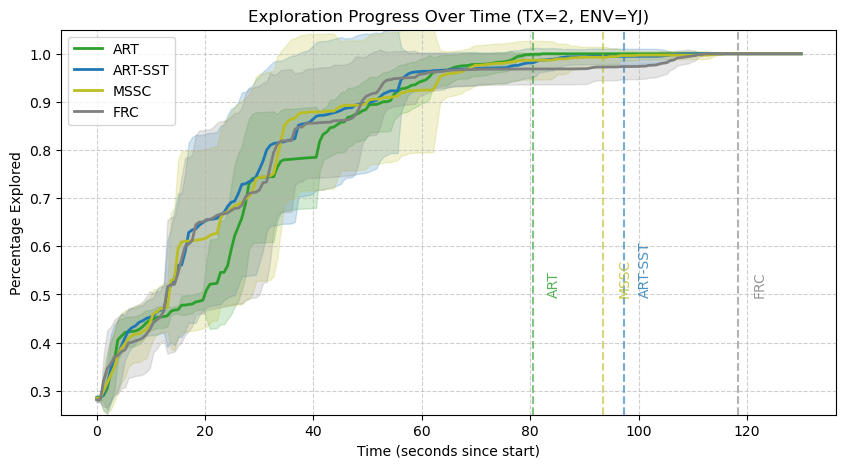}
    \includegraphics[width=1\linewidth]{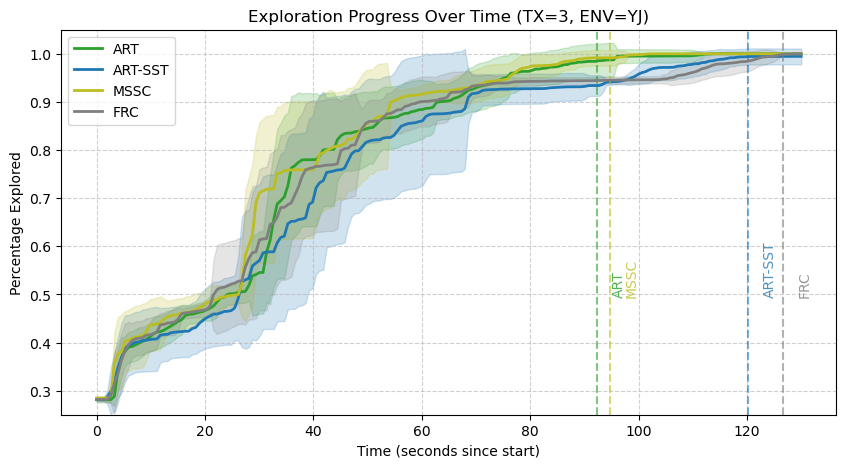}
    \caption{Percentage of environment explored over time in the Y-Junction environment. Vertical dashed lines indicate the mean time to full exploration. ART achieves the most efficient exploration across all payload sizes, with the most pronounced gains observed at 10 MB, where it completes exploration 23.7\% faster on average than competing methods. Even at the larger 100 MB payload, ART outpaces MSSC in completion time while requiring 4.5\% less travel distance. Complete numerical results for this environment are provided in Table \ref{table:combined-environments}.}
    \label{fig:yj-explore-tx2}
\end{figure}

\subsubsection{Environment: Window}

In the window environment, which models cave skylights and open archways that permit communication while restricting traversal, ART, ART-SST, and MSSC achieve faster exploration than FRC due to their ability to exploit available communication windows. ART consistently attains the shortest exploration times across nearly all payload sizes (Table~\ref{table:combined-environments}). For payloads ranging from 1~kB to 100~MB, ART selects transmission points that maximize throughput while minimizing backtracking. For the 10~MB payload, ART-SST produces a slightly shorter path, although ART maintains faster overall exploration due to more efficient data delivery. These results illustrate the trade-off between transmitting near the minimal usable signal threshold to reduce travel distance and reducing backtracking time at the cost of lower transmission rates.

\subsubsection{Environment: Y-Junction}
In the Y-Junction environment, ART, ART-SST, and MSSC exhibit similar performance for smaller payloads, as backtracking provides little advantage when a communication link is established. For payloads exceeding 100~kB, ART’s performance advantage increases (Fig.~\ref{fig:yj-explore-tx2}). Across all payload sizes, ART achieves the shortest path distances and fastest exploration times, with the largest improvement (23.7\%) occurring at the 10~MB payload. This effect arises from ART’s capacity to position the Scout optimally at the junction’s branching point, reducing round-trip costs. Increasing payload size amplifies ART’s relative benefit, indicating that adaptive transmission decisions become more valuable as the data payload size and/or transmission reliability requirements grow.

\subsubsection{Environment: Long Tunnel}
In the long tunnel environment the Scout and Specialist exit communication range. This results in extended transmission times for MSSC due to its reliance on reaching the nearest point with any detectable signal, regardless of link quality. FRC exhibits the longest exploration times and greatest distances traveled, reflecting the inefficiency of fixed rendezvous strategies in this large environment. ART achieves the shortest exploration times by balancing navigation backtracking with data transmission requirements.

ART consistently outperforms all baseline methods across payload sizes (Table~\ref{table:combined-environments}). The performance advantage is most pronounced for medium payloads (100~kB–10~MB), where ART maintains link quality while avoiding full returns to the Specialist. When strict signal thresholds are enforced (ART-SST) for the 100~MB payload, exploration efficiency approaches that of FRC, as the Scout must return nearly completely to the Specialist’s location.

MSSC demonstrates higher variability in exploration time across all payload sizes, indicated by elevated standard deviations. This variability reflects the sensitivity of minimal signal threshold strategies to fluctuations in environmental conditions and signal propagation. Because MSSC initiates transmission upon detecting the minimal usable signal, small changes in link quality or spatial configuration can substantially affect transmission points and overall mission duration. These results suggest that, while MSSC can occasionally produce efficient paths, it lacks the consistency required for reliable multi-robot coordination in communication-constrained environments.

\section{LIMITATIONS}
 While the three environments capture diverse subterranean geometries, they do not reflect the full complexity of real-world caves. Simulations assume a static noise floor and do not model electromagnetic interference which may arise in practical deployments. While our communication-aware planning strategy improves data exchange efficiency, it assumes the Scout backtracks to previously saved frontiers after transmission. Re-planning from the transmission location could further improve exploration efficiency, a promising direction for future work. Finally, all evaluations are performed in simulated subterranean environments. Validation in physical subterranean environments may provide greater evidence of real-world applicability, essentially substituting a more sophisticated backend for simulating signal attenuation.

\section{CONCLUSION}
Adaptive, communication-aware motion planning substantially enhances multi-agent exploration in communication-constrained environments. Within the Scout–Specialist paradigm, the Scout selects transmission points that jointly optimize navigation paths, signal quality, and data size, minimizing backtracking while maintaining reliable data delivery. The Adaptive RF-Transmission (ART) algorithm consistently outperforms baseline methods, achieving up to 51.7\% faster exploration and reducing path distance by up to 58.2\%, demonstrating the benefit of integrating motion planning with adaptive communication. The ART-SST variant illustrates how enforcing minimum signal thresholds can preserve high-fidelity data transfer, highlighting the trade-off between transmission reliability and exploration efficiency. These findings underscore the potential of communication-aware motion planning to enable efficient, coordinated multi-robot exploration in environments with limited connectivity.


\clearpage

\bibliographystyle{IEEEtran}

\bibliography{ref}

\end{document}